\documentclass{article}

\usepackage{ijcai22}

\usepackage{times}
\usepackage{soul}
\usepackage{url}
\usepackage[hidelinks]{hyperref}
\usepackage[utf8]{inputenc}
\usepackage[small]{caption}
\usepackage{graphicx}
\usepackage{amsmath}
\usepackage{amsthm}
\usepackage{booktabs}
\usepackage{algorithm}
\usepackage{algcompatible}

\title{Semi-Supervised Imitation Learning of Team Policies\\ from Suboptimal Demonstrations}

\iftrue
\author{
Sangwon Seo
\And
Vaibhav V. Unhelkar
\affiliations
Rice University
\emails
\{sangwon.seo, vaibhav.unhelkar\}@rice.edu
}
\fi

\usepackage{dsfont}
\usepackage{xspace}
\usepackage{bm}
\usepackage{bbold}
\usepackage{amsfonts}
\usepackage{multirow}
\usepackage{tikz}
\usetikzlibrary{arrows}
\usetikzlibrary{bayesnet}

\usepackage{subcaption}
\usepackage{enumitem}
\setlist[itemize]{leftmargin=*,noitemsep,topsep=2pt}
\setlist[enumerate]{leftmargin=*,noitemsep,topsep=2pt}

\usepackage{soul}

\newcommand{\boxdynamic}{\textit{Movers}\xspace}
\newcommand{\bagdynamic}{\textit{Cleanup}\xspace} 
 
\newcommand{\mail}{\textsc{MAIL}\xspace}
\newcommand{\btil}{\textsc{BTIL}\xspace}
\newcommand{\xhat}{\hat{x}} 
\newcommand{\mmdp}{MMDP\xspace}
\newcommand{\mdp}{MDP\xspace}
\DeclareMathOperator{\EX}{\ensuremath{\mathbb{E}}}
\newcommand\myeq{\mkern1.5mu{=}\mkern1.5mu} 
\newcommand\mycol{\mkern1.5mu{:}\mkern1.5mu} 
\newcommand\mysum{\mkern1.5mu{+}\mkern1.5mu}
\newcommand\mypm{\mkern1.5mu{\pm}\mkern1.5mu} 
\newcommand{\tableval}[2]{#1$\mypm$#2}
\newcommand{\tablevalbf}[2]{\textbf{#1}$\mypm$\textbf{#2}}
\newcommand{\jsd}{\textit{JS Div.}\xspace}
\newcommand{\hamming}{\textit{Hamming}\xspace}

\newcommand\blfootnote[1]{%
  \begingroup
  \renewcommand\thefootnote{}\footnote{#1}%
  \addtocounter{footnote}{-1}%
  \endgroup
}

\usepackage{soul}
\usepackage{balance}

\begin{document}
\maketitle
\blfootnote{This article is an extended version of an identically-titled paper accepted at the International Joint Conference on Artificial Intelligence (IJCAI) 2022.}

\newif\ifsupplementary
\supplementarytrue

\begin{abstract}
We present Bayesian Team Imitation Learner (\btil), an imitation learning algorithm to model the behavior of teams performing sequential tasks in Markovian domains.
In contrast to existing multi-agent imitation learning techniques, \btil explicitly models and infers the time-varying mental states of team members, thereby enabling learning of decentralized team policies from demonstrations of suboptimal teamwork.
Further, to allow for sample- and label-efficient policy learning from small datasets, \btil employs a Bayesian perspective and is capable of learning from semi-supervised demonstrations.
We demonstrate and benchmark the performance of \btil on synthetic multi-agent tasks as well as a novel dataset of human-agent teamwork. 
Our experiments show that \btil can successfully learn team policies from demonstrations despite the influence of team members' (time-varying and potentially misaligned) mental states on their behavior.
\end{abstract}

\section{Introduction}
\label{sec:introduction}
Teamwork is essential for the success of human enterprise.
As artificial agents increasingly become parts of human life, thus, they too are expected to reason about and contribute to human teams.
At the same time, teamwork is highly challenging to perfect.
Successful human teams employ a variety of training techniques to improve coordination and teamwork \cite{tannenbaum2020teams}.
Analogously, spurred by the need to enable and enhance human-agent collaboration, there has been growing work on developing computational techniques for training artificial agents to support human teams \cite{thomaz2016computational}.
These techniques build upon a variety of AI paradigms, such as planning under uncertainty, reinforcement learning, and imitation learning.

In this work, we consider the paradigm of imitation learning \cite{argall2009survey,osa2018algorithmic}, wherein an agent learns from demonstrations and (in contrast to reinforcement learning) can thereby learn policies for teamwork without the need of unsafe exploration.
By providing novel multi-agent imitation learning techniques that are inspired by real-world teaming considerations, this work aims to enable agents to model, assess, and improve both human-human and human-AI teamwork in sequential tasks.
Mathematically, imitation learning techniques seek to learn a single-agent behavioral policy $(\pi)$, a stochastic function that encodes probability of selecting an action $(a)$ in a task-specific context $(s)$, given a dataset of $(s, a)$-tuples provided by a demonstrator.
Typically, the context features $(s)$ and actions $(a)$ are assumed to be fully observable and measured using sensors.

Imitation learning has been extended to model multi-agent systems by seeking to learn a set of behavioral policies $\{ \pi_i | i\myeq1\mycol{n} \}$, one corresponding to each member $(i\myeq1\mycol{n})$ of the multi-agent system from demonstrations of teamwork \cite{le2017coordinated,bhattacharyya2019simulating,song2018multi,lin2019multi}.
We provide a brief survey of related multi-agent imitation learning (\mail) techniques in \ifsupplementary \autoref{sec:related-work}\else Appendix A\fi\@. \mail is an emerging area of research, wherein the existing works either focus on learning policies corresponding to game-theoretic equilibria of multi-agent systems, assume homogeneity in agents' capabilities, or assume data of optimal teaming behavior as the training input.
However, in contrast to most settings considered in prior art, teamwork observed in practice \cite{salas2018science,seo2021towards} often differs in three key ways: (1) it may not correspond to a game-theoretic equilibrium, (2) it can be suboptimal due to its dependence on latent performance-shaping factors (such as team members' mental models), and (3) it can involve team members with different capabilities. 

Informed by these three considerations of teamwork observed in the real world, in \autoref{sec:problem-statement}, we provide an alternate problem formulation of \mail for collaborative tasks.
In particular, we consider that the team members' behavior $(\pi_i)$ depends not only on the context features $(s)$ but also on their (time-varying) mental models $(x_i)$ pertaining to teamwork.
As the mental models cannot be readily sensed and manually annotating them is resource-intensive, they are modeled as partially observable to the imitation learner.
Further, in teaming scenarios where the mental models are misaligned (i.e., the team members do not maintain a shared understanding), the demonstrations available for learning can be suboptimal.
Thus, adding to the challenge, in our problem formulation the imitation learner also needs to infer which segments of the teaming demonstrations are (sub)-optimal and perform learning under partial observability of the dynamic state $(s,x)$.

\paragraph{Summary of Contributions.}
Towards this problem setting, we present Bayesian Team Imitation Learner (\btil, pronounced as ``bee-tul''), an imitation learning algorithm that can learn decentralized team policies from both optimal and sub-optimal demonstrations.
To effectively learn the team policies, \btil explicitly models the time-varying mental states of each team member $(x_i)$ and jointly learns their transition model $(T_x)$.
To enable sample- and label-efficient policy learning, \btil utilizes a Bayesian perspective and is capable of learning from partial supervision of the mental states.
We benchmark our solutions against two existing techniques \cite{pomerleau1991efficient,song2018multi}.
In our evaluations, we emphasize the challenge of collecting teaming demonstrations by collecting a novel dataset of human-agent teamwork for settings where (a) the labels of mental models are only partially available, and (b) the number of demonstrations is small relative to the size of the task's state space.
Our experiments show that \btil can learn team policies from small semi-supervised datasets of suboptimal teamwork and outperform the baselines across relevant metrics.

\begin{figure}[t]
  \centering
  \includegraphics[width=0.7\linewidth]{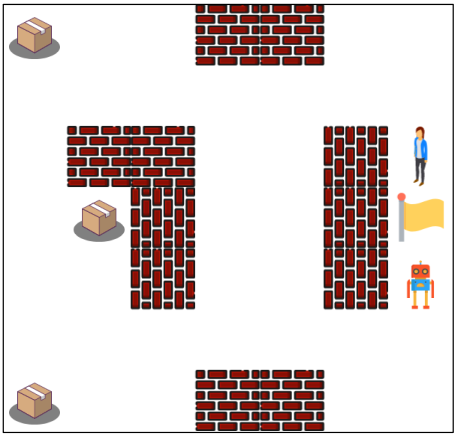}
  \caption{\boxdynamic~domain. The human-robot team needs to move the boxes to the flag. Each box can be moved if and only if both agents pick it up and move it along the same direction.}
  \label{fig.dynamic box}
\end{figure}

\paragraph{Running Example.}
To help describe our approach, we utilize the collaboration scenario shown in \autoref{fig.dynamic box} as a running example.
In this scenario, a two-member team composed of Alice and Rob is tasked with moving all boxes together to the flag in the least amount of time.
At each time step, each agent can choose to move in any one of four cardinal directions, attempt to pick up or drop a box, or perform no operation.
As each box is heavy, it cannot be lifted by one agent alone.
To effectively move all boxes to the goal location, the team needs to coordinate which box to pick next as well as the path along which to move the box.
Thus, in this scenario, the pertinent latent state $(x)$ corresponds to the team's next target location (i.e., one of the boxes or the flag), while the observable state $(s)$ corresponds to the locations of the agents and boxes.

During task execution, the team members' mental models $(x)$ can be misaligned.
For example, Alice may target the box at the top left while Rob targets the box at the bottom left, thereby demonstrating suboptimal teamwork.
Further, to improve the collaboration, team members may or may not choose to change their target at any point during the task based on behavior of the other teammate.
Thus, as motivated earlier, the demonstrations of teamwork may not correspond to a game-theoretic equilibrium and can be suboptimal due to dependence on partially observable, dynamic mental models.
The goal of the imitation learner is to learn the mental model-dependent policies of the team given these (potentially suboptimal) partially observable demonstrations of teamwork.

\section{Related Work}
\label{sec:related work summary}

Our work relates to the following three sub-areas of imitation learning: multi-agent imitation learning, learning from suboptimal demonstrations, and learning from partially observable demonstrations. 
Here, we summarize research from these sub-areas and relate it to our approach.
Please see \ifsupplementary \autoref{sec:related-work}\else Appendix A\fi\xspace for a more detailed discussion of  \mail techniques.

\paragraph{Multi-agent Imitation Learning (MAIL).}
Although multiple MAIL algorithms exist, the problem setting considered in prior art differs from the one considered in this paper.
Reiterating from \autoref{sec:introduction}, prior \mail techniques either learn behavior corresponding to game-theoretic equilibria of multi-agent systems \cite{song2018multi,lin2019multi}, assume homogeneity in agents' capabilities \cite{bhattacharyya2019simulating}, or do not consider latent performance-shaping factors (such as mental models or cognitive states).
Approaches that do model latent states assume that the latent state is either shared between members \cite{wang2021co,ivanovic2018generative} or time-invariant \cite{le2017coordinated}.
In contrast, inspired by real world teaming considerations, we seek to develop \mail algorithms that both recognize the individual and dynamically changing latent states of each member and are capable of learning multi-agent policies from different levels of supervision over the latent states.

\paragraph{Learning from Partially Observable Demonstrations.}
Imitation learning with occlusions or missing features has also received increasing attention in the last decade.
These techniques, while not directly valid for the multi-agent
setting considered in this work, inform our work. 
\cite{torabi2018behavioral,sun2019adversarial} consider learning an agent policy from demonstrations that may not include data of the demonstrator's actions.
Similarly, \cite{choi2011inverse,gangwani2020learning} allow incomplete specification of states by utilizing a belief state.
\cite{unhelkar2019learning} explicitly model an agent's latent decision factors (e.g., mental states), which can change dynamically within an episode.
While related to our work, these techniques only consider single-agent tasks and do not model the interaction between multiple agents.
\cite{bogert2018multi} provide an approach for multi-robot inverse reinforcement learning from partially observable demonstrations.
In contrast to our approach, their work does not model agents' mental states or seek to learn from demonstrations of suboptimal teamwork.

\paragraph{Learning from Suboptimal Demonstrations.}
While classical imitation learning assumes demonstrations are generated from experts who behave optimally, a few approaches admit that demonstrations can be suboptimal in practice.
For example, with an assumption that the majority of demonstrations are optimal, \cite{choi2019robust,zheng2014robust} focus on imitation learning that is robust to suboptimal outliers.
Meanwhile, \cite{brown2019extrapolating,chen2021learning,zhang2021confidence} aim to incorporate demonstrations that come from demonstrators  whose level of expertise is unknown in order to overcome the challenge of scarce expert demonstrations.
\cite{yang2021trail} utilize a latent action representation while learning the optimal policy from potentially suboptimal demonstrations. 
While related to our approach, these solutions for imitation learning from suboptimal demonstrations neither consider demonstrators' mental states nor multi-agent tasks.
In contrast, our goal is to develop an approach that can learn stochastic multi-agent policies from demonstrations that are \textit{both} {partially observable} and {suboptimal}.

\section{Problem Formulation}
\label{sec:problem-formulation}
To formalize the problem of learning team policies from suboptimal and partially observable demonstrations, we first provide models for the team task and team members' behavior.

\subsection{Task Model}
\label{sec:task-model}
Due to our focus on learning task-oriented team policies, we require a model to represent team tasks.
Borrowing from prior research in multi-agent systems \cite{oliehoek2016concise}, we build upon the framework of multi-agent Markov decision processes (\mmdp) to describe the tasks of interest.
An \mmdp models sequential collaborative tasks and is specified by the tuple $ M_{\text{task}} \doteq (n, S, A, T, R, \gamma) $, where
\begin{itemize}
\item $n$, is the number of agents $i$ indexed $1:n$;
\item $s \in S$, denotes the set of task states;
\item $a_i \in A_i$, is the set of actions $a_i$ available to the $i$-th agent;
\item $A\myeq\times_i A_i$ is the set of joint actions, where $a\myeq[a_1, \cdots, a_n]$ denotes the joint action;
\item $T_s(s'|s, a): S \times A \times S \rightarrow [0, 1]$ denotes the state transition probabilities, i.e., the probability of the next task state being $s'$ after the team agents executed action $a$ in state $s$
\item $R(s, a): S \times A \rightarrow \mathbb{R}$ is the joint reward that the team receives after execution action $a$ in state $s$.
\item $\gamma$ is the discount factor.
\end{itemize}

The \mmdp model assumes that all agents have a shared objective and each agent has full observability of the task state and reward.
The shared objective of the set of $n$ agents, whom we jointly refer to as the \textit{team}, is to maximize their expected cumulative discounted reward, $\EX[\Sigma_t \gamma^t R(s_t, a_t)]$.
In this work, we focus on team tasks that can be modeled as \mmdp where the set of states $S$ and the set of actions $A$ are finite. 
Several real-world tasks can be modeled using \mmdp. 
For instance, the scenario described in the running example can be described as an \mmdp with $n\myeq2$, $S$ modeling the task-relevant features (namely, the agent and box locations), and $A$ modeling the actions available to the agent.

The solution to the \mmdp is a set of $n$ decentralized agent policies $\pi_{1:n}$, where $\pi_i$ is the policy of the $i$-th agent.
In the running example, this corresponds to policies of Alice and Rob.
Mathematically, $\pi_i(a_i|s)$ is a probability distribution of the $i$-th agent’s actions $a_i$ conditioned on the \mmdp state.
Since each agent has full state observability, in theory, an \mmdp can be solved optimally in a centralized manner by the team using \mdp solvers \cite{puterman1990markov} before the task begins.
If each team member follows this optimal policy faithfully, coordination between team members in \mmdp tasks is guaranteed during task execution.

\subsection{Agent Model}
\label{sec:agent-model}
In practice, however, one seldom observes perfect coordination among team members, including in tasks where the agents have complete or near-complete observability of the task state and complete knowledge of the team's objective (e.g., healthcare team in an operating room, or a team of basketball or soccer players).
The potential causes of this imperfect coordination are varied.
For instance, imperfect coordination can occur due to inability to compute a joint policy, lack of prior coordination, imperfect execution, and different individual preference.
To design an imitation learning algorithm that can effectively recover team policies, it is essential to explicitly consider these imperfections and latent causes of suboptimal teamwork.

Hence, to model teamwork observed in practice, we provide a latent variable model for each team member’s (potentially suboptimal) behavior.
Our model extends the Agent Markov Model (AMM), which explicitly models latent states of a single agent's behavior \cite{unhelkar2019learning}, to model teamwork.
In particular, we model each team member's behavior as the tuple $(X_i, b_{x_i}, T_{x_i}, \pi_{i})$, where

\begin{itemize}
\item $x_i \in X_i$ denotes the latent states that influence the $i$-th agent's behavior during the task. These may include mental models, methods to tie-break if multiple optimal policies exist, or preferences over different task components. 
\item $b_{x_i}(x_i) \in X_i  \rightarrow [0, 1]$ denotes the probability distribution of the latent state at the start of the task.
\item  $T_{x_i}(x_i'|s,x_i,a,s') \in  S \times X_i \times A \times S \times X_i  \rightarrow [0, 1]$ denotes the transition model of the latent state.
\item $\pi_{i}(a_i | s, x_i) \in S \times X_i \times A_i \rightarrow [0, 1]$ denotes the team member's policy, a probability distribution of each member's decision $a_i$ conditioned on their decision factors $(s,x_i)$.
\end{itemize}
Referring to the running example, behavior of Alice and Rob depends not only the task context (\mmdp state) but also on their latent preferences over the next target location.
For each team member, the agent model helps in modeling this latent preference (as $x_i \in X_i$), their latent state-dependent policy (as $\pi_i$), and the evolution of their latent preference (via $b_{x_i}$ and $T_{x_i}$).
While the above model is expressive and can represent a variety of team behaviors (e.g., suboptimal policies, evolution of latent preferences based on past behavior), we assume the transition dynamics $T_x$ to be Markovian for computational tractability.
Notationally, we jointly refer to the behavioral models for the whole team as $(X, b_x, T_x, \pi)$, where $X\myeq \times_i X_i$, $T_x \myeq \{T_{x_1}, \cdots, T_{x_n}\}$, and $\pi \myeq [\pi_1, \cdots, \pi_n]$.
The latent state of whole team is denoted as $x\myeq[x_1, \cdots, x_n]$.

\subsection{Problem Statement}
\label{sec:problem-statement}
In the classical \mail setting, the goal is to learn the team policy from a set of teamwork demonstrations%
\footnote{\textbf{Notation}: We use superscript to denote the time step. The subscript is overloaded and, based on the context, is used to denote the $i$-th agent, $m$-th demonstration, task state $s$, agent's latent state $x$, or action $a$. Finally, for notational convenience, we use $\mathds{1}_{abc}(a', b', c')$ to denote $\mathds{1}(a'\myeq a, b'\myeq b, c'\myeq c)$.} %
$\tau \doteq (s^{0:h}, a^{0:h})$, where $h$ denotes the demonstration length.
In our setting, however, team behavior is additionally influenced by the trajectories of team members' latent states $\chi \doteq (x^{0:h})$, which are partially observable and resource-intensive to annotate.
Hence, we focus on semi-supervised learning of team policies, where $x$-labels are available for only a subset of the demonstrations.

Formally, our problem corresponds to learning the team policy $\pi$, given the \mmdp task model $(n, S, A, T, R, \gamma)$, a set of $d$ observable demonstrations, $\tau_{1:d} \doteq \{\tau_m\}_{m=1}^d$, and labels of $x$ for a subset $l(\leq d)$ of the demonstrations, $\chi_{1:l} \doteq \{ \chi_m \}_{m=1}^{l}$.
For our running example, the problem corresponds to recovering the behavioral policy of the two-agent team (Alice and Rob), given $d$ observable trajectories of agent and box locations $\tau_{1:d}$ and labels of each agents' preferred target locations for a subset of the trajectories $\chi_{1:l}$.

\section{Solution: Static Latent States}

For ease of exposition, we first derive the policy learning algorithm for the case wherein team members do not change their latent mental model $(x)$ during task execution; mathematically, $T_x \doteq \mathds{1}(x\myeq x')$.
In the next section, we build upon the solution derived for this special case to solve the overall problem of \autoref{sec:problem-statement}.
We note that, despite the simplification of static latent states, the learner needs to reason under partial state observability to learn the team policy.

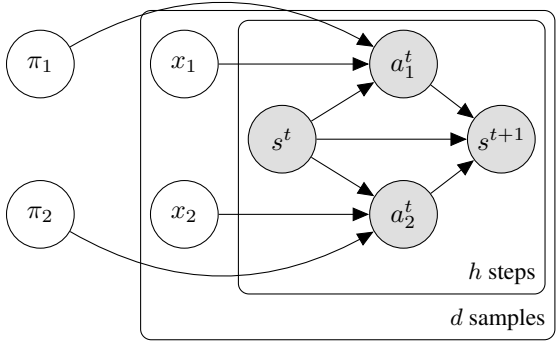
\begin{figure}
    \centering
    \scalebox{1}{
        \begin{tikzpicture}
            \node[obs, minimum size=0.9cm] (st) {$s^t$} ;
            \node[obs, right=2cm of st, minimum size=0.9cm] (st1) {$s^{t+1}$} ;
            \node[obs, right=0.7cm of st, yshift=1cm, minimum size=0.9cm] (a1) {$a_1^t$} ;
            \node[obs, right=0.7cm of st, yshift=-1cm, minimum size=0.9cm] (a2) {$a_2^t$} ;
    
            \node[latent, left=2cm of a1, minimum size=0.9cm] (x1) {$x_1$} ;
            \node[latent, left=2cm of a2, minimum size=0.9cm] (x2) {$x_2$} ;
            \node[latent, left=1cm of x1, minimum size=0.9cm] (pi1) {$\pi_1$} ;
            \node[latent, left=1cm of x2, minimum size=0.9cm] (pi2) {$\pi_2$} ;
    
            \edge {st} {a1} ;
            \edge {st} {a2} ;
            \edge {a1} {st1} ;
            \edge {a2} {st1} ;
            \edge {st} {st1} ;
            \edge {x1} {a1} ;
            \edge {x2} {a2} ;
            \path (pi1) edge [->, bend left] (a1) ;
            \path (pi2) edge [->, bend right] (a2) ;
            \plate {h} {(a1)(a2)(st)(st1)} {$h$~steps} ;
            \plate {M} {(h)(x1)(x2)} {$d$~samples} ;
    
        \end{tikzpicture}
    }
    \caption{Dynamic Bayesian network for the behavior of a 2-agent team with time-invariant latent states depicted using plate notation.}
    \label{fig. static}
\end{figure}
\paragraph{Generative Model.}
\newcommand{\LSLS}{\chi_{1:l}}  
\newcommand{\USLS}{\{x_m\}_{m>l}}  
\newcommand{\JP}{\pi}  

To enable policy learning from a small number of demonstrations, we utilize a Bayesian approach and provide a generative model of team behavior.
The generative model, shown in \autoref{fig. static} for a two-agent team, models the process of generating team demonstrations.
Each agent (indicated by the subscript) selects its action, $a_i^t \sim \pi_i(\cdot | s^t, x_i)$, at time step $t$, based on the current task state $s^t$, her mental state $x_i$, and policy $\pi_i$.
In this special case, the latent state $x_i$ does not change during task execution and, thus, is represented without a superscript.
The dynamics of task state $s^{t+1}$ depend on the \mmdp model $T_s(\cdot|s^t, a^t)$.

To complete the generative process, the model additionally includes priors (not shown in \autoref{fig. static}) for the latent state and policy.
In absence of any additional domain knowledge, we assume that the policy is given as a Categorical distribution.
Hence, we define the prior of a policy as its conjugate prior, Dirichlet distribution: $\pi_{i, sx} \sim \text{Dir}({u}^{\pi})$, where ${u}^{\pi}\myeq(u_1^{\pi}, \cdots, u_{|A|}^{\pi})$ are hyperparameters.
Similarly, we assume that the latent state is drawn from the uniform distribution unless additional information is given: $x_i \sim \text{Uni}(X)$.
Given the generative model and data of semi-supervised demonstrations $(\tau_{1:d}, \LSLS)$, the policy can be learned by maximizing the posterior: $ p(\JP | \tau_{1:d}, \LSLS) \label{eq.pi_posterior}$.

\paragraph{Policy Learning with Supervision $(l\myeq{d})$.}
For the case when labels of $x$ are available for the entire dataset, we can directly compute the posterior distribution of the policy as:
\begin{align}
    p(\pi_{i, sx} | \tau_{1:d}, \LSLS ) 
    &= \text{Dir}({w}_{i, sx}) 
\end{align}
where, $w_{i, sxa} = {u}^{\pi}_a + \sum_{m=1}^l \sum_{(s', a')\in\tau_m}\mathds{1}_{sxa}(s', x_{i, m}, a_i')$.

\paragraph{Policy Learning with Semi-supervision $(l<d)$.}
\label{sec: static policy learning}
When labels of the latent state are only partially available, the likelihood $p(\tau_{1:d}, \LSLS|\JP)$ cannot be readily computed as it depends on unknown variables, namely, the subset of the data for which latent states $x$ labels are unavailable: $\USLS$. 
Hence, to calculate the posterior distribution in a computational tractable manner, we explore paradigms for approximate Bayesian computation.
We note the computing the posterior through exact inference (i.e., brute-force) is intractable due to the high-dimensional nature of our problem.
Inspired by prior work on single-agent modeling \cite{johnson2014stochastic,unhelkar2019learning}, we utilize mean-field variational inference (MFVI) \cite{beal2003variational} and derive an MFVI algorithm for modeling team policies.
In MFVI, the posterior of the team policy is approximated as the variational distribution $q(\JP)$ that maximizes the evidence lower bound (ELBO).
For our problem, the ELBO corresponds to:
\begin{align}
    \mathcal{L}(q) \doteq \EX_q \left[ \log \frac{p(\JP, \USLS, \text{data})}{q\left(\JP\right)q\left(\USLS\right)} \right] \label{eq.elbo}
\end{align}
The solution for the optimization problem, $\arg\max_q \mathcal{L}(q)$, corresponds to the iterative computation of the local $q(x)$ and global $q(\JP)$ variational distributions until convergence:
\begin{align}
    q(\pi_{i, sx}) &= \text{Dir}\left({w}_{i, sx}^\pi \right) \label{eq.pi_approx_post}\\
    q(x_{i, m} = x) &= (1/Z) \exp\left[ \textstyle \sum_{\tau_k} \ln \tilde{\pi}_{i, sxa} \right] \label{eq.x_approx_dist}
\end{align}
where, $Z=\sum_{x'\in X} \exp\left[ \sum_{\tau} \ln \tilde{\pi}_{i, sx'a} \right]$ is the partition function, the operator $\tilde{A}$ denotes $\exp\left(\EX_{q'(A)}\left[\ln A \right]\right)$, and
\begin{align}
    w_{i, sxa}^\pi &= u_a^\pi +
    \sum_{k=1}^m \EX_{q'(x_{i, m})}\sum_{\tau_m}\mathds{1}_{sxa}(s^t, x_{i, m}, a_i^t). \label{eq.hyperparam_w}
\end{align}%
Given the posterior $q(\pi)$, the policy is simply estimated as the maximum a posteriori (MAP) estimate: $\hat{\pi} = \arg\max_\pi q(\pi)$.
\section{Solution: Dynamic Latent States}
\label{sec: solution-extensions}
In contrast to the special case considered above, in practice, each member's task-specific mental model may evolve during the task.
Hence, next, we extend the solution presented for static latent states to solve the general problem of \autoref{sec:problem-statement}.
Analogous to the previous section, we first provide a generative model of team behavior and provide a MFVI-based approach for computing the MAP estimate of policy.

\paragraph{Generative Model.}
\newcommand{\ldls}{\chi_{1:l}}  
\newcommand{\udls}{\{x_{m}^{0:h}\}_{m>l}}  
\newcommand{\JT}{T_x}  

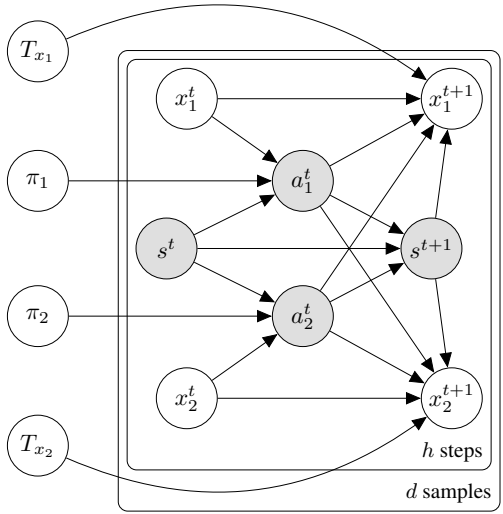
\begin{figure}[t]
    \centering
    \scalebox{0.9}{
    \begin{tikzpicture}
        \node[obs, minimum size=0.9cm] (st) {$s^t$} ;
        \node[obs, right=3cm of st, minimum size=0.9cm] (stn) {$s^{t+1}$} ;
        \node[obs, right=1.1cm of st, yshift=-1cm, minimum size=0.9cm] (a2) {$a_2^t$} ;
        \node[obs, right=1.1cm of st, yshift=1cm, minimum size=0.9cm] (a1) {$a_1^t$} ;

        \node[latent, below=1.3cm of st, xshift=0.3cm, minimum size=0.9cm] (x2) {$x_2^t$} ;
        \node[latent, above=1.3cm of st, xshift=0.3cm, minimum size=0.9cm] (x1) {$x_1^t$} ;
        \node[latent, below=1.3cm of stn, xshift=0.3cm, minimum size=0.9cm] (x2n) {$x_2^{t+1}$} ;
        \node[latent, above=1.3cm of stn, xshift=0.3cm, minimum size=0.9cm] (x1n) {$x_1^{t+1}$} ;

        \node[latent, left=3cm of a2, minimum size=0.9cm] (pi2) {$\pi_2$} ;
        \node[latent, left=3cm of a1, minimum size=0.9cm] (pi1) {$\pi_1$} ;
        \node[latent, below=1.5cm of pi2, yshift=0.5cm, minimum size=0.9cm] (t2) {$T_{x_2}$} ;
        \node[latent, above=1.5cm of pi1, yshift=-0.5cm, minimum size=0.9cm] (t1) {$T_{x_1}$} ;

        \edge {st} {a2} ;
        \edge {st} {a1} ;
        \edge {a2} {stn} ;
        \edge {a1} {stn} ;
        \edge {st} {stn} ;
        \edge {x2} {a2} ;
        \edge {x1} {a1} ;
        \edge {pi2} {a2} ;
        \edge {pi1} {a1} ;
        \edge {x2} {x2n} ;
        \edge {x1} {x1n} ;
        \edge {a2} {x2n} ;
        \edge {a1} {x1n} ;
        \edge {a1} {x2n} ;
        \edge {a2} {x1n} ;
        \edge {stn} {x2n} ;
        \edge {stn} {x1n} ;
        \path (t2) edge [->, bend right] (x2n) ;
        \path (t1) edge [->, bend left](x1n) ;
        \plate {h} {(a2)(a1)(st)(stn)(x2n)(x1n)} {$h$~steps} ;
        \plate {M} {(h)(x2)(x1)} {$d$~samples} ;

    \end{tikzpicture}
    }
    \caption{Dynamic Bayesian network for the behavior of a 2-agent team with time-varying latent states depicted using plate notation.} 
    \label{fig.graph_model}
\end{figure}

For the general setting, the generative process of team behavior additionally needs to model the temporal evolution of the team members' mental models.
Thus, as shown in \autoref{fig.graph_model}, we augment the generative model of \autoref{fig. static} to include each agent's time-varying latent states $x_i^t$ and their dynamics: $T_{x_i} \doteq P(x_i^{t+1}|s^t, x_i^t, a^t, s^{t+1})$.
In general, how the mental model evolves may not be known a priori and, thus, requires specification of a prior distribution.
Similarly to the policy prior, we utilize Dirichlet distribution as the prior for the latent state transition model, i.e, $T_{x_i, sxas'} \sim \text{Dir}(u_1^T, \cdots, u_{|X|}^T)$, where $u_x^T$ are hyperparameters.
Both $T_x$ and $\pi$ jointly influence the likelihood of the semi-supervised demonstrations of team behavior, $p(\tau_{1:d}, \ldls | T_x, \pi)$, where $T_x$ specifies the distribution of the next latent state $x^{t+1}$ and $\pi$ that of the action $a^t$.
Due to this dependence, to recover the team policy using MFVI, we need an approach to compute their joint posterior $p(T_x, \pi | \tau_{1:d}, \ldls )$.

\begin{algorithm}[tb]
\caption{Bayesian Team Imitation Learner (\btil)}
\label{alg.policy_learning}
\textbf{Input}: $\tau_{1:d}, \ldls$ \\
\textbf{Parameters}: $u^{\pi}, u^{T_x}, N, T_s$
\begin{algorithmic}[1] 
\STATE Initialize $w_i^\pi, w_i^{T_x}$ for $i\myeq 1:n$
\STATE Initialize posterior of all unlabeled states $q(\udls)$
\WHILE{$\mathcal{L}(\mathrm{q})$ converges}
\STATE Update the variational parameters $w_{1:n}^{\pi}, w_{1:n}^{T_x}$
\FORALL{$\tau_m$} \label{algline.qx_start}
\STATE Compute forward $F$ and backward $B$ messages
\STATE \resizebox{0.91\linewidth}{!}{Update posterior of all unlabeled states $q(\udls)$}
\ENDFOR
\ENDWHILE
\STATE Compute the policy posterior $q(\pi) \sim \text{Dir}(w_i^\pi)$
\STATE \textbf{return} $\arg\max_{\pi} q(\pi)$
\end{algorithmic}
\end{algorithm}

\paragraph{Bayesian Team Imitation Learner (\btil).}
Similar to \autoref{sec: static policy learning}, our solution to the overall problem -- abbreviated as \btil -- includes iterative computation of variational distributions.
However, in contrast to the previous section, the variational distributions additionally include the posterior of latent state dynamics $T_x$.
The \btil algorithm builds upon MFVI and is derived by maximizing the following ELBO:
\begin{align}
    \mathcal{L}(q) := \EX_q \left[ \log \frac{p\left(\JP, \JT, \udls, \text{data}\right)}{q\left(\JP \right)q\left(\JT \right)q\left(\udls\right)} \right] \label{eq.elbo2}
\end{align}

\autoref{alg.policy_learning} provides the pseudocode of \btil, which approximates the posterior distribution $p(T_x, \pi | \tau_{1:d}, \ldls )$ as independent variational distributions $q(\pi) q(T_x)$, where $q(\pi_i)\myeq \text{Dir}({w}_i^\pi)$ and $q(T_{x_i}) \myeq\text{Dir}(w_i^{T_x})$.
The estimates of posterior distributions are improved by iteratively updating the variational parameters $w^\pi, w^{T_x}$ (line 4).
Similar to \autoref{eq.hyperparam_w}, the variational parameters are updated as:
\begin{align}
    w_{i, jkass'}^{T_x} &\myeq u_k^{T_x} \mysum \sum_{m\myeq1}^d \EX_{q(x)} \left[ \textstyle\sum_{t}\mathds{1}_{jkass'}(x_i^{t:t+1}, a^t, s_i^{t:t+1}) \right]  \nonumber\\
    w_{i, sxa}^\pi &= u_a^{\pi} \mysum \sum_{m\myeq1}^d \EX_{q(x)} \left[ \textstyle \sum_{t}\mathds{1}_{sxa}(s^t, x_{i, m}^t, a_i^t) \right] \label{pi_param} 
\end{align}

An estimate of the posterior distribution of unlabeled states $q(x)$ is required to compute the expectations in \autoref{pi_param}.
This local variational distribution is given as follows:
\begin{align}
q(x_{i, m}^{0:h}) &\propto \exp\left( \EX\left[ \ln p(x_i, \text{data}|T_{x_i}, b_{x_i}, \pi_i, T_s \right]\right) \nonumber\\
&= p(x_{i, m}^{0:h}, \tau_{i, m}|\tilde{T}_{x_i}, \tilde{\pi}_i, T_s, b_{x_i}) / Z_{i, m} \label{eq. local dist}
\end{align}
To compute \autoref{eq. local dist} in a tractable manner, we define the following forward-backward messages.
These messages are computed in a recursive manner (line 6) as follows:
\begin{align*}
    F(t, j_1, &\cdots, j_n) \\
        &\doteq P(x_1^t\myeq{}j_1, \cdots, x_n^t\myeq{}j_n, s^{0:t}, a^{0:t})\\
        &\myeq{} \sum_{k_1, \cdots, k_n} F(t-1, k_1, \cdots, k_n) T_s \prod_{i\myeq{}1}^n \left(\tilde{T}_{x_i} \tilde{\pi}_i\right)\\
    B(t, j_1, &\cdots, j_n) \\
        &\doteq P( s^{t+1:h}, a^{t+1:h}|x_1^t\myeq{}j_1, \cdots, x_n^t\myeq{}j_n,s^{0:t}, a^{0:t})\\
        &= \sum_{l_1, \cdots, l_n} B(t+1, l_1, \cdots, l_n) T_s \prod_{i\myeq{}1}^n \left(\tilde{T}_{x_i} \tilde{\pi}_i \right) \\
    F(0, j_1, &\cdots, j_n) = \prod_{i=1}^n\tilde{b}_{x_i} \tilde{\pi_i}, \quad 
    B(h, j_1, \cdots, j_n) = 1
\end{align*}
For notational convenience, the subscript $m$ is omitted. Given the forward $F$ and backward messages $B$, \btil computes the required local probabilities (line 7) as follows:

\begin{align*}
    q(x^t) &\propto F \cdot B \\
    q(x_i^t) &= \sum_{{x}_{-i}}q(x_1^t, \cdots, x_n^t) \\
    q(x^t, x^{t+1}) &\propto F\cdot \prod_i\left(\tilde{T}_{x_i}\tilde{\pi}_i\right)\cdot T_s\cdot B\\
    q(x_i^t, x_i^{t+1}) &= \sum_{{x}_{-i}}q(x^t, x^{t+1})
\end{align*}
The time complexity of forward and backward messaging passing subroutine is $O(h|X|^{2n})$.
Given the converged posterior $q(\pi)$ in line 10, the team policy is estimated as the MAP estimate (line 11).

\begin{table*}[ht]
\small
\centering
\begin{tabular}{llcccccccc}
\toprule
                                  &                 & \multicolumn{4}{c}{\boxdynamic}                                                                       & \multicolumn{4}{c}{\bagdynamic}                          \\
                                                    \cmidrule(lr){3-6}                                                                                      \cmidrule(lr){7-10}
                                  &                 & \multicolumn{2}{c}{Alice}                        & \multicolumn{2}{c}{Rob}                        & \multicolumn{2}{c}{Alice}                          & \multicolumn{2}{c}{Rob} \\
                                                    \cmidrule(lr){3-4}                                  \cmidrule(lr){5-6}                                  \cmidrule(lr){7-8}                                    \cmidrule(lr){9-10}
                                Setting  &     Algorithm            & \hamming                 & \jsd                      & \hamming                 & \jsd                      & \hamming                   & \jsd                      & \hamming                 & \jsd     \\
\midrule
\multirow{5}{*}{\textit{with $T_x$}} & \textit{Random }     & \tableval{0.29}{0.00}   & \tableval{0.14}{0.00}   & \tableval{0.29}{0.00}   & \tableval{0.14}{0.00}   & \tableval{0.18}{0.00}   & \tableval{0.32}{0.00}   & \tableval{0.18}{0.00} & \tableval{0.36}{0.00}  \\
                                  & \textit{BC}     & \tableval{0.32}{0.02}   & \tableval{0.15}{0.01}   & \tableval{0.31}{0.03}   & \tableval{0.15}{0.01}   & \tablevalbf{0.06}{0.01}   & \tableval{0.22}{0.03}   & \tablevalbf{0.03}{0.02} & \tableval{0.17}{0.03}  \\
                                  & \textit{MAGAIL} & \tableval{0.30}{0.03}   & \tableval{0.21}{0.02}   & \tableval{0.31}{0.03}   & \tableval{0.24}{0.01}   & \tableval{0.16}{0.03}     & \tableval{0.29}{0.02}   & \tableval{0.17}{0.05}   & \tableval{0.38}{0.04}  \\
                                  & \textit{\btil-Sup}     & \tableval{0.30}{0.01}   & \tableval{0.07}{0.00}   & \tableval{0.27}{0.01}   &  \tableval{0.07}{0.00}  & \tableval{0.09}{0.02}     & \tableval{0.20}{0.01}   & \tableval{0.04}{0.01}   & \tableval{0.19}{0.00}  \\   \cmidrule{2-10}

                                  & \textit{\btil-Semi}   & \tablevalbf{0.15}{0.02} & \tablevalbf{0.05}{0.00} & \tablevalbf{0.16}{0.02} & \tablevalbf{0.05}{0.00} & \tableval{0.09}{0.02}     & \tablevalbf{0.07}{0.01} & \tableval{0.04}{0.01}   & \tablevalbf{0.04}{0.00}  \\
\midrule
\multirow{3}{*}{\textit{w/o $T_x$}}  & \textit{Random}     & \tableval{0.72}{0.00}   & \tableval{0.14}{0.00}   & \tableval{0.77}{0.00}   & \tableval{0.14}{0.00}   & \tableval{0.78}{0.00}   & \tableval{0.32}{0.00}   & \tableval{0.82}{0.00} & \tableval{0.36}{0.00}  \\
                                  & \textit{\btil-Sup}     & \tableval{0.31}{0.01}   & \tableval{0.07}{0.00}   & \tableval{0.35}{0.01}   & \tableval{0.07}{0.00}   & \tableval{0.54}{0.01}     & \tableval{0.20}{0.01}   & \tableval{0.48}{0.03}   & \tableval{0.19}{0.00}  \\ \cmidrule{2-10}
                                  & \textit{\btil-Semi}   & \tablevalbf{0.30}{0.01} & \tablevalbf{0.04}{0.00} & \tablevalbf{0.33}{0.01} & \tablevalbf{0.04}{0.00} & \tablevalbf{0.42}{0.01}   & \tablevalbf{0.05}{0.00} & \tablevalbf{0.36}{0.01} & \tablevalbf{0.04}{0.00} \\
\bottomrule
\end{tabular}
\caption{Results on the synthetic data of multi-agent teamwork averaged over five learning trials.}
\label{table.dynamic result}
\end{table*}
\section{Experiments}
\label{sec:numerical-expt}

We evaluate \btil using two domains, \boxdynamic and \bagdynamic, which include aforementioned features and challenges of real world teamwork.
These domains build upon the \textit{cooperative box pushing} task of \cite{oliehoek2016concise}, include unambiguous latent preferences, and allow for opportunities of (mis)-alignment of team members' mental models.
Video demonstrations of collaborative task execution in these domains are included in the supplementary material.

Due to the latent nature of mental models, labels of human team members' latent state cannot be ascertained without significant manual effort and annotation.
Further, existing multi-agent dataset, to our knowledge, have not recorded mental models at each time step of collaborative tasks. 
Hence, to conduct the proposed experiments and compute pertinent metrics, we create two novel datasets for each domain: one synthetically generated and the other collected via human subject experimentation.

\subsection{Domains}
\paragraph{\boxdynamic.}
This domain realizes the running example of \autoref{sec:introduction} in a $7\times7$ grid world. 
The two member team of Alice and Rob is tasked with carrying boxes to the goal position (flag).
Boxes cannot be picked up by one agent alone; hence, to efficiently complete the task, the agents should coordinate on their latent preference over which box to pick or drop next.
Each box can be either on its original location, held by both agents, or on the goal location.
As described in the running example, agents can take one of the six actions at each step.
In this domain, there are $38988$ observable states and five possible mental states (corresponding to the three box pickup locations and two drop off locations), resulting in around $200000$ states affecting each team member's decisions.

\paragraph{\bagdynamic.}
This domain has a similar configuration as \boxdynamic, but the environment includes lighter trash bags are placed instead of heavy boxes.
Trash bags can only be picked up by one agent; hence, to effectively complete this task, each agent should carry different trash bags as much as possible.
Each trash bag can be either on its original location, held by one of the agents, or be dropped at the goal location. 
In this domain, there are in total over $450000$ states affecting agent decisions.
The set of primitive actions and latent preferences available to each agent are same as the \boxdynamic domain.

\subsection{Baselines and Metrics}
\label{sec: baseline}
Due to the novel features of our problem setting, to the best of our knowledge, existing algorithms do not readily apply to the general version of our problem.
Existing MAIL solutions either do not model mental states, assume them to be aligned across all team members, or model them as time invariant.
Hence, we benchmark our approach against baselines on special cases of our problem.
In our design of experiments, we divide our problem into four settings based on two criteria: (a) whether the transition model of latent states $T_x$ is known a priori or not, and (b) whether the latent states are completely labeled or not. 
We apply the behavioral cloning (BC) and MAGAIL as baselines for the setting of complete labels and the known transition model $T_x$ \cite{pomerleau1991efficient,song2018multi}.
The implementation of BC and MAGAIL is similar to those used in \cite{ho2016generative,song2018multi} respectively but adapted to handle discrete states.
For the settings where labels are only partially available, we cannot apply existing algorithms; instead, we compare the performance of the supervised (\btil-Sup) and the semi-supervised (\btil-Semi) version of our approach.
Implementation details of \btil and the baselines are provided in \ifsupplementary \autoref{sec: implementation}\else Appendix D\fi\xspace.

We evaluate our approach's ability to effectively learn team policies using the weighted Jensen-Shannon divergence (\jsd) between the true and learned policies.
Like \cite{unhelkar2019learning}, the policy divergence metric is weighted by the relative counts of states $(s, x)$ observed in the training set. 
The policy learning performance can only be computed in experiments with synthetic data, where the ground truth policy is known.
Hence, in addition, we compare the ability to decode the unlabeled latent states using learned policies.
In particular, we utilize the normalized Hamming distance (\hamming) between the decoded $\hat{x}_{i, m}^{0:h}$ and true $x_{i, m}^{0:h}$ sequences of latent states as the decoding metric.
To decode the team member's latent states, as detailed in \ifsupplementary \autoref{sec:decoding}\else Appendix B\fi\xspace, we extend the algorithm presented by \cite{seo2021towards}.
Lastly, to have a better sense of the worst case values of these highly nonlinear metrics, we utilize the Random baseline, which models the team policy as a Uniform distribution.

\subsection{Results on Synthetic Data}
\label{sec:synthetic-data-results}

We first present results on the synthetic dataset, which are summarized in \autoref{table.dynamic result}.
These experiments evaluate \btil in two settings: with and without prior knowledge of $T_x$.
We present additional results in \ifsupplementary \autoref{sec: additional-results}\else Appendix E\fi\xspace.

\paragraph{Data of Multi-agent Teamwork.} 
The first dataset is synthetically generated by simulating teamwork between two artificial agents.
For each domain, we implement the Markovian task model $T_s$, specify ground truth policies $\pi_i$, and transitions of the team members $T_{x_i}$.
To arrive at the agent policies, we specify rewards associated with each latent state, utilize value iteration to compute $Q$-values, and derive stochastic policies $\pi_i$ using the softmax operation over $Q$-values.
Execution sequences are created by first assigning initial latent states $x_i$ to each team member and, then iteratively, (a) sampling team members' action $a_i \sim \pi(\cdot | s, x_i)$, (b) sampling the next state $s' \sim T(\cdot | s, a)$, and (c) sampling the next latent state $x_i' \sim T_{x_i}(\cdot |x_i, s, a, s')$ until the task termination criteria or $200$ time steps are reached. 
For each domain, we generate $200$ demonstrations for training and $100$ for evaluation.
The proportion of suboptimal training demonstrations (as defined in \ifsupplementary \autoref{sec:user-study}\else Appendix C\fi\xspace) is 49\% and 7\% for \boxdynamic and \bagdynamic, respectively.

\paragraph{\btil Outperforms Baselines in Fully Supervised Settings.}
The first four rows of \autoref{table.dynamic result} provide results with $20$ labeled demonstrations $({d}\myeq{l}\myeq{20})$ and $T_x$ as inputs.
For this setting in the \boxdynamic domain, we observe that the supervised version of our algorithm (\btil-Sup) learns more accurate team policies than the baselines.
Other baselines (BC, MAGAIL) performed no better than the Random baseline.
In \bagdynamic, our algorithm also outperformed other baselines in the policy learning metric (\jsd). Despite a small training set, \btil can learn the team policy with low JS-Divergence by effectively leveraging the $x$-labels and knowledge of $T_x$.

In terms of the decoding metric (\hamming), BC showed better results than \btil in \bagdynamic domain.
We posit that this trend occurs due to a combination of two reasons. 
First, the \boxdynamic domain requires tighter coordination relative to the \bagdynamic domain.
In \boxdynamic, teammates must agree on which object to pick next to achieve coordination while in \bagdynamic, they only need to ensure that they are not picking the same object next.
Second, the decoding metric (\hamming) assesses learning performance only on a subset of the state space (i.e., the states encountered in the test set); while \jsd assess learning performance for the entire state space and, thus, is a better indicator of generalizability.
This explains why, even in \bagdynamic, BC outperforms \btil only on the decoding metric.
In the \boxdynamic domain, which requires tighter coordination, BC performs poorly in both metrics.

\paragraph{\btil Is Capable of Learning Team Policies Without Prior Knowledge of $T_x$.}
As shown in the bottom section of \autoref{table.dynamic result}, \btil can maintain its policy learning performance (\jsd) even when $T_x$ is unknown.
These results suggest that \btil is capable of learning $T_x$ along with $\pi$ and that joint learning of $(\pi, T_x)$ is essential to imitation learning of team policies.
Somewhat unsurprisingly, when the decoding algorithm utilizes the learnt $T_x$, the latent state decoding performance degrades relative to the known $T_x$ case.

\paragraph{\btil Effectively Utilizes Unsupervised Demonstrations to Improve Team Policy Learning.}
Lastly, we compare the performance of \btil under semi-supervision, i.e., the general setting of \autoref{sec:problem-statement}.
For these trials (denoted as \btil-Sup), we provide the algorithm additional $180$ demonstrations without latent state labels $(d\myeq{200},l\myeq{20})$.
Comparing performance of \btil-Sup and \btil-Semi, we observe improvement in policy learning performance, highlighting the ability of \btil to effectively leverage available unsupervised data.
This ability is particular critical in practice, where collecting data of $(s,a)$-tuples can be significantly less resource intensive than arriving at the labels of mental states.
In additional experiments reported in \ifsupplementary \autoref{sec: additional-results}\else Appendix E\fi\xspace, which further investigate the effect of training set size and amount of semi-supervision on the learning performance, we observe that the decoding performance is enhanced with more labeled data and semi-supervision provides the most benefit when the labeled training set is small.

\subsection{Results on Data of Human-AI Teamwork}
\label{sec:user-study-results}

Synthetically generated data, while being useful in validating policy learning performance, cannot capture the variability in behavior demonstrated by humans and human-agent teams.
Hence, to benchmark our approach in more realistic settings, we evaluate our algorithm on a novel dataset of human-agent teamwork through an ethics board-approved human subject experiment with $33$ participants ($16$ female, $17$ male, mean age: $26.7 \pm 5.3$ years), who were recruited at Rice University.

\paragraph{Data Collection Procedure.}
To collect this novel dataset of human-AI teamwork, we designed a web-based interface shown in \autoref{fig.web}.
Through this web-based interface participants completed tasks in the two domains (\boxdynamic and \bagdynamic) with an AI teammate.
The human participant served the role of Alice, while the AI teammate served the role of Rob depicted as a robot avatar.

When the experiment starts, participants are asked to complete a short demographic survey. 
Further, before the participants conduct any tasks, they were provided with an interactive tutorial to make them familiar with the task and the interface.
Upon completing the interactive tutorial, the experiment consisted of $9$ sessions: $4$ sessions on the \boxdynamic~ domains and $5$ tasks for \bagdynamic.
For each domain, the first two sessions were used as practice sessions.
These practice sessions were designed for participants to gain further familiarity with the user interface.
Through the tutorials and practice sessions, each participant was requested to complete tasks collaboratively with a robot avatar which behaves according to an AI policy.
The AI policy is generated similar to the synthetic experiment, i.e, by specifying rewards associated with each mental model, running value iteration, and taking the softmax function over $Q$-values.
\definecolor{clrgreen}{HTML}{D9EEE1}
\definecolor{clrwhite}{HTML}{FFFFFF}
\definecolor{clrgrey}{HTML}{CCCCCC}
\definecolor{clrdarkgrey}{HTML}{888A89}
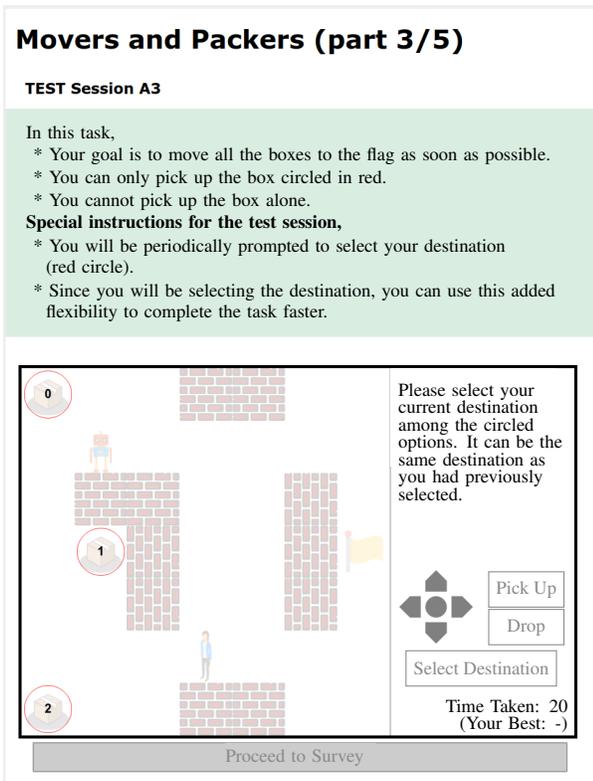
\begin{figure}[t]
  \centering
  \begin{tikzpicture}[font=\scriptsize]
    \draw (0, 0) node[inner sep=0] {\includegraphics[width=0.92\linewidth]{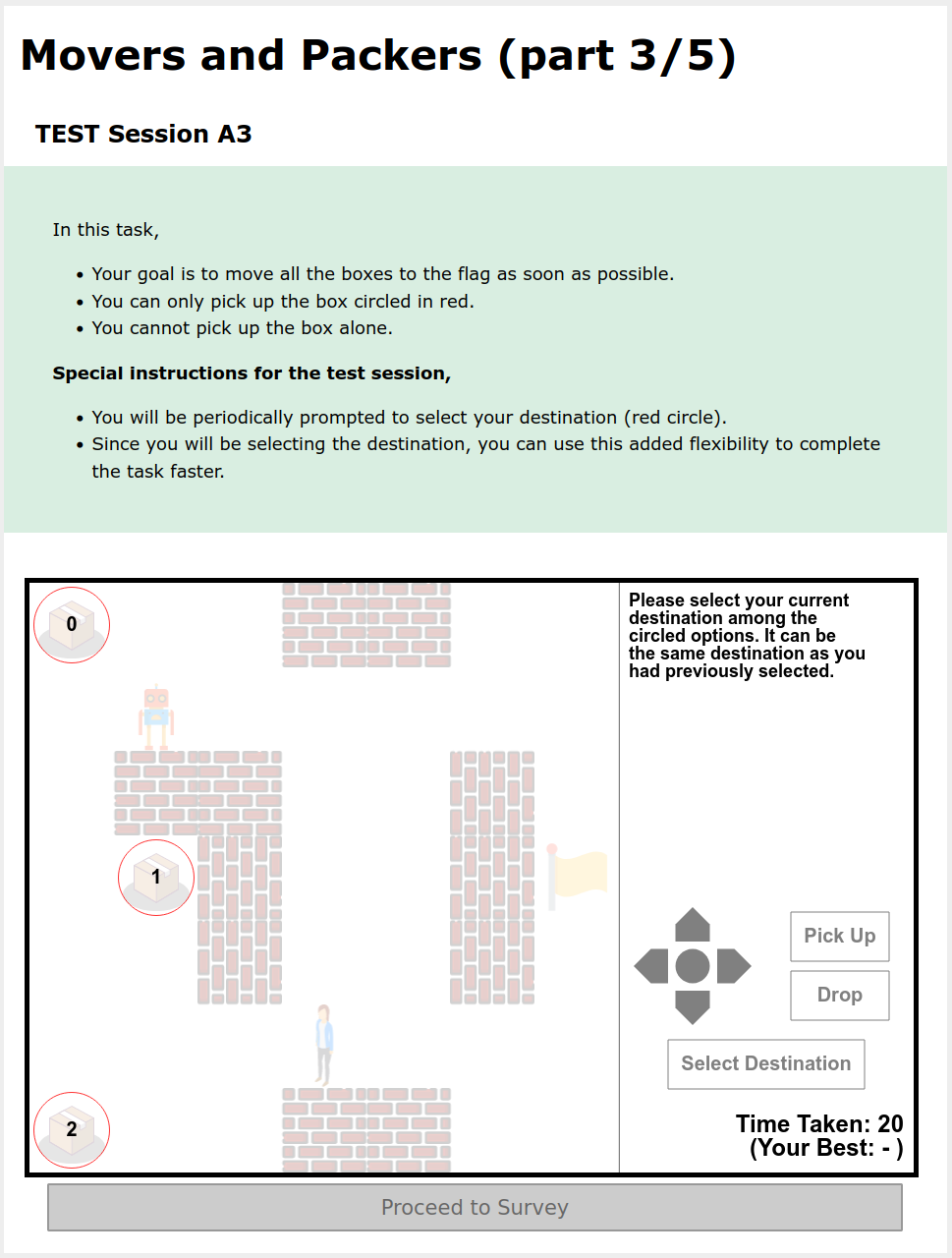}};
    \fill [fill=clrgreen] (-3.7,1) rectangle ++(7.5,2.5);
    \draw (0, 3.5) node[minimum width=6cm, align=left, text width=7.3cm] {In this task,};
    \draw (0, 3.2) node[minimum width=6cm, align=left, text width=7.1cm] {* Your goal is to move all the boxes to the flag as soon as possible.};
    \draw (0, 2.9) node[minimum width=6cm, align=left, text width=7.1cm] {* You can only pick up the box circled in red.};
    \draw (0, 2.6) node[minimum width=6cm, align=left, text width=7.1cm] {* You cannot pick up the box alone.};
    \draw (0, 2.3) node[minimum width=6cm, align=left, text width=7.3cm, font={\bfseries\scriptsize}] {Special instructions for the test session,};
    \draw (0, 1.85) node[minimum width=6cm, align=left, text width=7.1cm] {* You will be periodically prompted to select your destination \\ ~~(red circle).};
    \draw (0, 1.25) node[minimum width=6cm, align=left, text width=7.1cm] {* Since you will be selecting the destination, you can use this added \\ ~~flexibility to complete the task faster.};
    \fill [fill=clrwhite] (1.2, -0.925) rectangle ++(2.2,1.3);
    \draw (2.4, -0.6) node[minimum width=2cm, align=left, text width=2.2cm] {Please select your\\[-0.2em] current destination \\[-0.2em] among the circled \\[-0.2em] options. It can be the \\[-0.2em] same destination as \\[-0.2em] you had previously \\[-0.2em] selected.};
    \fill [fill=clrgrey] (-1, -4.9) rectangle ++(2,0.29);
    \draw (0, -4.8) node[minimum width=1cm, text=clrdarkgrey, align=left, text width=2cm,font=\scriptsize] {Proceed to Survey};
    \draw [draw=clrdarkgrey, fill=clrwhite] (2.5,-2.8) rectangle node[text=clrdarkgrey, font=\scriptsize]{Pick Up} ++(1,0.47);
    \draw [draw=clrdarkgrey, fill=clrwhite] (2.5,-3.3) rectangle node[text=clrdarkgrey, font=\scriptsize]{Drop} ++(1,0.47);
    \draw [draw=clrdarkgrey, fill=clrwhite] (1.4,-3.85) rectangle node[text=clrdarkgrey, font=\scriptsize]{Select Destination} ++(2,0.47);
    \fill [fill=clrwhite] (1.5,-4.5) rectangle node[font=\scriptsize,minimum width=1.5cm, align=right, text width=2cm]{Time Taken: 20\\[-0.2em] (Your Best: -)} ++(2.1,0.5);
  \end{tikzpicture}
  
  \caption{A still of the user interface designed and used for collecting data of human-agent teamwork. Please also see the supplementary material for video demonstrations of the user interface.}
  \label{fig.web}
\end{figure}

For each task, participants' actions (clicking action buttons) are logged along with the task state (location of Alice, Rob, and the boxes) and their chosen destination (mental state corresponding to which box to pick up or drop next).
In general, collecting ground truth values of participants' mental states is challenging; in these experiments, we achieve this through a destination-selection interface detailed in \ifsupplementary \autoref{sec:user-study}\else Ap-pendix C\fi\@. By encouraging them to complete the task as soon as possible through the provided instructions and by showing ``Your Best" score, we gamify the collaborative task and expect participants to take goal-oriented actions.

Through this experiment, we collect $66$ trajectories for \boxdynamic, and $99$ trajectories for \bagdynamic, of which every time step is labeled. The average lengths of trajectories (in terms of time steps) are 68.1 in \boxdynamic and 40.1 in \bagdynamic. The proportions of suboptimal demonstrations are 17\% and 7\% in \boxdynamic and \bagdynamic, respectively. We use two thirds of the collected trajectories for training and the rest for evaluation.

\begin{table}
  \small
  \centering
  \begin{tabular}{lccc}
    \toprule
Algorith &  Supervision & \boxdynamic & \bagdynamic \\
    \midrule
    Random & N/A & \tableval{0.72}{0.00} & \tableval{0.77}{0.03} \\
    \btil-Sup & \textit{$100$\%} & \tableval{0.14}{0.02} &  \tableval{0.30}{0.02}    \\
    \btil-Semi & \textit{$50$\%} & \tableval{0.16}{0.02} &   \tableval{0.35}{0.03}    \\
    \btil-Semi & \textit{$20$\%} & \tableval{0.20}{0.02} &   \tableval{0.42}{0.02}    \\
    \bottomrule
  \end{tabular}
  \caption{State decoding performance (Hamming distance) on the data of human-agent teamwork averaged over five learning trials.}
  \label{table.inference human}
\end{table}

\paragraph{Performance of \btil Observed with Data of Multi-agent Teamwork Translates to Learning Policies of Human-agent Teamwork.}
As we cannot ascertain the true policy of a human, we utilize only the state decoding metric in these experiments.
Further, as would be the case in practice, the mental model dynamics are also unavailable; hence, $T_x$ needs to be learned by our algorithm and the MAGAIL baseline cannot be applied.
\autoref{table.inference human} summarizes the decoding performance computed using $\pi$ and $T_x$ learned by variants of \btil.
All variants of \btil are supplied with same amount of $(s,a)$-demonstrations but varying amount of $x$-labels.
\btil, even with a small amount of supervision, significantly outperform the Random baseline. 
In conjunction with the results computed with synthetic data, these experiments provide proof-of-concept in the ability of \btil to learn team policies from small semi-supervised datasets of optimal and suboptimal teamwork.

\section{Concluding Remarks}
\label{sec:discussions}

We provide \btil, a Bayesian approach to learn team policies from demonstrations of suboptimal teamwork.
In most collaboration scenarios, it is challenging to collect large labeled datasets of teamwork due to changes in team membership, adaptations in team policies, and the need for labeling latent states.
Inspired by these and other aspects of teamwork observed in practice, \btil includes multiple desirable features, including (a) the ability to learn from small sets of semi-supervised data, (b) explicit modeling of team members' mental models and model alignment and (c) the ability to jointly infer team policy, latent state dynamics, and latent states.
We confirm the ability of our algorithm to learn team policies on two novel datasets of teamwork, including one of human-AI teamwork.
Our work also offers several avenues of future work, including the ability to consider collaborative tasks where the task state $s$ itself may be partially observable and consideration of communicative actions.

\section*{Acknowledgments}
We thank the anonymous reviewers for their detailed and constructive feedback.
Sangwon Seo was partially supported by the Army Research Office through Cooperative Agreement Number W911NF-20-2-0214.
The views and conclusions contained in this document are those of the authors and should not be interpreted as representing the official policies, either expressed or implied, of the Army Research Office or the U.S. Government.
The U.S. Government is authorized to reproduce and distribute reprints for Government purposes notwithstanding any copyright notation herein.

\appendix
\setcounter{table}{0}
\renewcommand{\thetable}{\Alph{section}-\arabic{table}}
\setcounter{figure}{0}
\renewcommand{\thefigure}{\Alph{section}-\arabic{figure}}
\section{Brief Review of Multi-Agent Imitation Learning (MAIL)}
\label{sec:related-work}

While \mail is an emerging area, spurred by advances in deep learning and generative modeling, multiple algorithms have been developed for learning multi-agent policies from demonstrations in the last five years.
For instance, \cite{song2018multi,liu2020multi} and \cite{yang2020bayesian} extend generative adversarial imitation learning (GAIL) of \cite{ho2016generative} to multi-agent settings.
Similar to GAIL, these multi-agent GAIL techniques match the state and action occupancy between the learner and demonstrator policy through a generative adversarial approach.
These approaches focus on inferring the multi-agent policies at game-theoretic equilibrium; this learning of equilibrium policy is achieved by constraining the multi-agent reward to correspond to one of the game-theoretic equilibria of the multi-agent task.
These algorithms have been demonstrated to be successful in learning team policies from expert demonstrations; however, in many real-world scenarios, teamwork policies need to be learned from a mixture of expert and non-expert demonstrations.
Our work targets this alternate setting and seeks to learn team policies from both optimal and sub-optimal demonstrations.

\cite{bhattacharyya2018multi,bhattacharyya2019simulating} seek to learn a single agent policy which is shared among all agents.
Their technique is particularly useful for multi-agent systems with homogenous agents.
In contrast, our approach (\btil) does not require agents to be homogeneous.
As such, \btil can be applied to hybrid human-agent teams (e.g., as shown in the experiments) as well as human teams which include team members with different expertise (such as a team of doctors and nurses \cite{seo2021towards}).

In contrast to directly learning team policies, as in the aforementioned approaches as well as \btil, inverse reinforcement learning (IRL) has also been utilized to learn multi-agent policies.
These multi-agent IRL techniques first recover reward and then compute policies given reward through planning-based or reinforcement learning methods.
However, to ensure computational tractability of the reward learning step, multi-agent IRL techniques typically require strong assumptions regarding the structure of the shared reward.
For example, the multi-agent IRL technique of \cite{natarajan2010multi} models the joint reward as weighted sum of individual agent's reward and assumes that the state-space can be decomposed into agent-wise states.
\cite{yu2019multi} assumes demonstrations are generated under the logistic stochastic best response equilibrium.
Further, \cite{lin2017multiagent,wang2018competitive,lin2019multi} specifically tackles two-agent Markov game problems.

More importantly, all approaches discussed thus far do not explicitly consider latent decision factors, thereby making it challenging to consider tasks where multiple optimal policies exist or policies depend on team members' mental states.
We note that if ground-truth labels of latent variables are available (i.e., the fully supervised case of problem presented in \autoref{sec:problem-formulation}), these approaches can be readily extended to model latent states as we do so in our experiments.
However, if labels of certain decision factors are unavailable or partially available, these approaches are inapplicable. 

Recognizing the importance of latent decision factors on team behavior, recently, a few approaches have been developed that explicitly model latent states during multi-agent imitation learning.
For instance, the work of \cite{le2017coordinated} focuses on learning the latent state corresponding to the role of each member in a team task.
However, they model the relation between the latent feature and observable features using a hidden Markov model, and do not consider the effect of other agents' actions on the transition of the latent features.
Our problem assumes the role of each team member to be known a priori.
In future work, exploring the intersection of our approach and that of \cite{le2017coordinated} remains an area of interest.

\cite{wang2021co} extend the single-agent InfoGAIL algorithm due to \cite{li2017infogail} to learn multiple joint policies that may vary according to the latent feature (strategy code), particularly for the two-agent (human-robot) collaborative setting.
However, they assume the latent feature is shared between members (i.e., mental models of team members are aligned), time-invariant, and provide an unsupervised learning approach.
In contrast, our work is capable of considering tasks where team member's latent preferences may change during the tasks and learning from semi-supervised data.
\cite{ivanovic2018generative} utilize a Conditional Variational Autoencoder (CVAE) which also models latent space to learn multi-agent policies.
However, similarly to the work of \cite{wang2021co}, their approach does not allow for the value of latent feature to differ among agents.

\section{Decoding Team Members' Latent States}
\label{sec:decoding}

For effective human-agent teamwork, inferring latent states of other team members by artificial agents is critical \cite{thomaz2016computational}. 
Further, more recently, algorithms for decoding latent states of teamwork have also been proposed to assess quality of teamwork \cite{seo2021towards}. 
Hence, we utilize latent state decoding as an alternate metric to benchmark the performance of our solutions.
In this section, we describe the problem of latent state decoding in collaborative multi-agent settings as well as algorithms to solve this problem.

\subsection{Problem statement}
The goal of latent state decoding is to infer the mental models of team members on a test set, given a learned model of team behavior.
More concretely, the problem consists of following inputs, outputs, and parameters:
\begin{itemize}
    \item \textbf{Input}: a demonstration of teamwork, i.e., a trajectory $\tau$ of observable states $s^{0:h}$ and team actions $a^{0:h}$
    \item \textbf{Output}: decoded value of each team member's latent states $x_1^{0:h}, \cdots, x_n^{0:h}$ for the test demonstration
    \item \textbf{Parameters}: $(S, A, T_s, T_x, \pi)$
\end{itemize}
The decoding algorithm assumes both the team policy $\pi$ and latent state dynamics $T_x$ as known parameters.
While computing the decoding performance of our algorithm and the baselines, we utilize the learned values of $\pi$.
Further, for the problem settings where $T_x$ is known, we utilize the known values of $T_x$ to compute the decoding performance.
In contrast, for the problem settings where $T_x$ is unknown, we utilize its learned value to compute the decoding performance.

\subsection{Decoding Algorithm}
Similar to policy learning, we estimate the values of unknown latent states using a Bayesian approach.
In particular, we compute the MAP estimate for latent state of each member:
\begin{align*}
    \xhat_i^{0:h} &= \arg\max_{x_i^{0:h}}p(x_i^{0:h}|\tau) \\
    &= \arg\max_{x_i^{0:h}}p(x_i^{0:h}, s^{0:h}, a^{0:h})
\end{align*}
To solve this equation, we define the intermediate term $V$ and compute it recursively via dynamic programming:
\small
\begin{align*}
    V_t(r) \doteq& \max_{x_i^{0:t-1}}p(x_i^t \myeq{} r, x_i^{0:t-1}, s^{0:t}, \mathbf{a}^{0:t})\nonumber\\
    \myeq{}& \begin{cases}
                \pi_i(a_i^t|x_i^t \myeq{} r, s^t)\cdot \\
                \max_m \left[ T_{x_i}(x_i^t\myeq{}r|x_i^{t-1}\myeq{}m, s^{t-1:t}, a^{t-1}) \cdot  V_{t-1}(m)  \right]\\
             \end{cases} \\
    V_0(r) \myeq{}& \pi_i(a_i^0|x_i^0\myeq{}r, s^0)b(x_i^0\myeq{}r)
\end{align*}
\normalsize
Given team policy $\pi$ and transition $(T_s, T_x)$ specifications, the algorithm computes $V_0, \cdots, V_k$ using dynamic programming and finds the path that maximizes $V_k$ by backtracking to $t=0$ to find the decoded latent state sequence.

\section{Data Collection of Human-AI Teamwork}
\label{sec:user-study}
In this section, we provide additional details of the human subject experiment introduced in \autoref{sec:user-study-results} to collect data of human-AI teamwork.

\paragraph{Ground Truth of Human Mental States}
As noted in \autoref{sec:user-study-results}, the web-based user interface is used to collect data of participants' actions, task states, and mental states.
We reiterate that the mental states in the experimental domains correspond to the participants' intent regarding which box or bag to pick up or drop next.
While the actions and task states are observable and can be readily logged through the interface, collecting ground truth values of participants' latent mental states is challenging as participants can not only choose any of the items (boxes in \boxdynamic and bags in \bagdynamic) to pick up or drop next but also change their preference during task execution during the test sessions.
Hence, to reliably collect the ground truth of human mental states, we design and utilize a destination-selection interface.

A demonstration of the interface is provided in the supplementary video attachments.
The destination-selection interface, when activated, overlays the task environment with a translucent layer, disables participants' action buttons, and only permits the participants to select their current destination (i.e., mental state).
The destination-selection interface is introduced during the interactive tutorial, which is provided before the data collection procedure, to make participants familiar with its use and features.
Further, to ensure alignment between participants' mental state and the reported destination (i.e., estimate of the ground truth used for learning experiments), the user interface includes several features: (a) the participants are periodically prompted to select their destination, (b) the selected destination is marked using a red circle, and (c) the participants are allowed to pick up and drop a box / bag only at the marked destination.

In particular, the destination-selection interface is activated if either five steps have passed since the previous time it was activated or if the state of any box / bag changes (e.g., if one of the agents picks or drops any box / bag).
Additionally, participants are also provided with a ``Select Destination'' button using which they can manually activate the destination-selection interface to report their mental state any time during the experiment.
During the data collection, on average, the destination-selection interface was invoked $14.9$ times per episode in \boxdynamic domain and $9.7$ times per episode in \bagdynamic domain.
For comaprison, the average trajectory length (in terms of time steps)  is $68.1$ in \boxdynamic domain and $40.1$ in \bagdynamic domain.

\paragraph{Suboptimal Demonstrations}
In this work, we refer to those team demonstrations as suboptimal where the team exhibits poor coordination due to misaligned between team members' mental models.
In \boxdynamic domain, this definition of suboptimality translates to timesteps when the mental states of Alice and Rob are different.
For instance, if Alice and Rob are targeting different boxes to pick up, or if Alice wants to drop the box to its original location while Rob tries to drop it to the flag, we consider the state-action pair of that time step as suboptimal.
In \bagdynamic domain, we consider a state-action pair as suboptimal if one of the following conditions hold: there are more than one bag on the floor but both agents try to pick up the same bag; there is at least one bag on the floor but an agent is targeting the bag that the other agent is currently holding; or an agent who is currently holding a bag tries to drop it on its original location.
Please note that if Alice is moving the last bag towards the flag, the state-action pair is considered optimal, regardless of Rob's mental state.

\begin{figure*}[ht]
     \centering
     \begin{subfigure}[b]{0.45\textwidth}
         \centering
         \includegraphics[width=\textwidth]{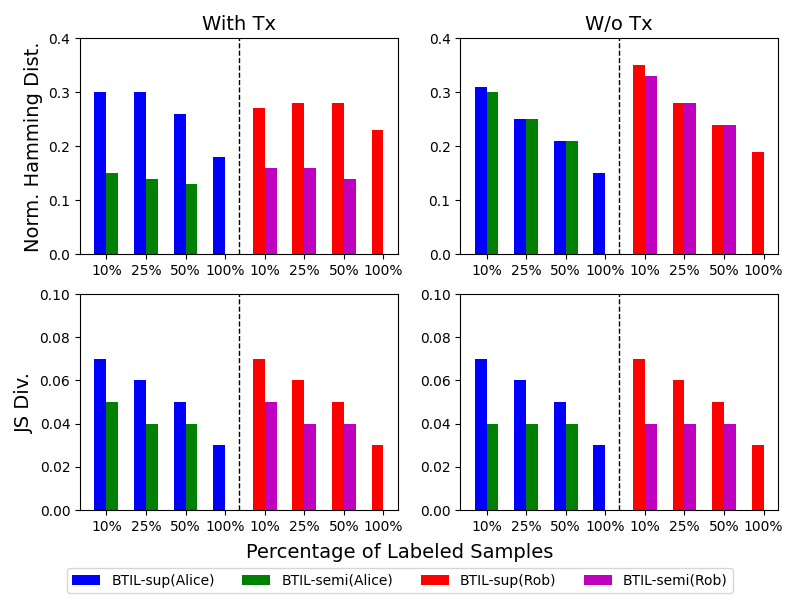}
         \caption{\boxdynamic}
         \label{fig:box results}
     \end{subfigure}
     \begin{subfigure}[b]{0.45\textwidth}
         \centering
         \includegraphics[width=\textwidth]{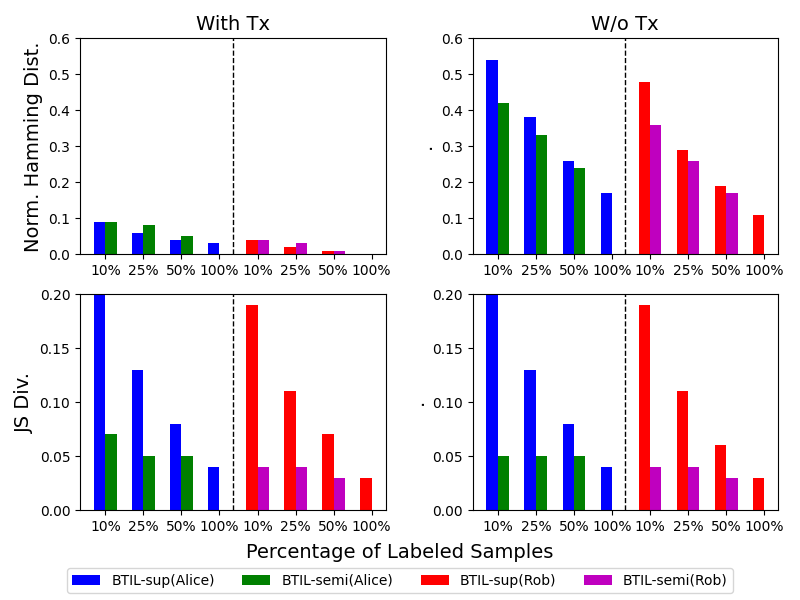}
         \caption{\bagdynamic}
         \label{fig:bag results}
     \end{subfigure}
    \caption{Effect of training set size $(d)$ and amount of semi-supervision $(l/d)$ on the policy learning (\jsd) and state decoding (\hamming) performance of \btil. Results computed using the synthetic data of multi-agent teamwork.}
    \label{fig. supervision}
\end{figure*}

\section{Experiments: Implementation Details}
\label{sec: implementation}

\paragraph{Behavioral Cloning (BC)}
For BC, we grouped state $s$ and action $a_i$ pairs of each agent by its latent state $x_i$.
Then, we applied BC to each group.
Namely, we trained $|X|$ policies corresponding to each latent state value for each agent.
The policy network and value network consisted of three-layered perceptrons with $64$, $128$, and $128$ nodes in each layer.

\paragraph{Multi-Agent Generative Adversarial Imitation Learning (MAGAIL)}
For MAGAIL, because our domain is cooperative, we utilized the centralized version of \cite{song2018multi}, and used PPO for its RL procedure, informed by the recent paper \cite{wang2021co}.
Each policy generator network took $(s, x_i)$ as its input and outputted $a_i$, while the value network took $(s, x_1, x_2)$ as its input in a centralized manner.
Policy generator networks and the value networks of MAGAIL consisted of three-layered perceptrons with $64$, $128$, and $128$ nodes in each layer., identical to BC.
The discriminator of MAGAIL was implemented with two-layer perceptrons with $128$ nodes in each layer. 
Since our domain consists of discrete states $(s, x_1, x_2)$ and actions $(a_1, a_2)$, we one-hot encoded each $s, x_1, x_2, a_1, a_1$ and fed them to each network. 
We trained each algorithm for $500$ iterations with a batch size of $128$.
Before running MAGAIL, we pretrained its policy network for $100$ iterations.
Then, for the first ten iterations, we updated the discriminator $100$ times per each policy update, and then reduced the update number to five times per each policy update.
We used the Adam optimizer \cite{kingma2014adam} with a fixed learning rate of $7$e$-4$.

\paragraph{Bayesian Team Imitation Learner (\btil)}
We set the hyperparameters of Dirichlet distributions to $u^\pi= 1.2$ and $u^{T_x} = 1.01$ for \boxdynamic, and to $u^\pi= 1.01$ and $u^{T_x} = 1.01$ for \bagdynamic.
Full specification of $T_{x_i}(x_i'|s, x_i, a, s')$ is a $S \times X \times A \times S \times X$ tensor.
For each agent, the number of elements to learn $T_{x_i}$ reaches $1.37\cdot10^{12}$ and $7.69\cdot 10^{12}$ for each domain.
This requires around 5 TiB and 28 TiB to store the tensors as float type, respectively.
Hence, we approximate the transition as $T_{x_i}(x_i'|s, x_i, a)$ by ignoring the dependence on the next state $s'$. 

\section{Effect of Training Set Size and Level of Supervision on Learning Performance}
\label{sec: additional-results}

Lastly, we investigate the effect of training set size $(d)$ and amount of semi-supervision $(l/d)$ on the policy learning (\jsd) and state decoding (\hamming) performance of \btil. 
\autoref{fig. supervision} demonstrates the performance enhancement according to the level of supervision.
With more labeled data, we observe that performance improves on both metrics.
Augmenting unlabeled data further improved the performance for all settings except \bagdynamic with $T_x$.
In this setting, we reason that the decoding performance was already high that augmented unlabeled samples did not provide any further information.
In general, the effect of augmented samples is the most dramatic when we use fewer number (e.g., 10\%) of labeled samples. 

Further, in addition to \autoref{table.dynamic result}, we additionally investigated the performance of our approach along with baselines using a larger supervised dataset $(d=l=200)$.
\autoref{table. 200 samples} summarizes the results computed using the synthetic data of multi-agent teamwork.
Similar trends observed hold; however, due to the availability of a larger dataset, all algorithms offer improved performance.

\begin{table}[ht]
\small
\centering
\begin{tabular}{lllccc}
\toprule
                                    &                 &         & BC   & MAGAIL & \btil\\
\midrule
\multirow{4}{*}{\boxdynamic} & \multirow{2}{*}{Alice} & \hamming & 0.23 & 0.22 & 0.18 \\
                                    &                        & \jsd   & 0.06 & 0.15  & 0.03 \\
\cmidrule(r){2-3}
                                    & \multirow{2}{*}{Rob}   & \hamming & 0.26 & 0.20  & 0.23 \\
                                    &                        & \jsd   & 0.06 & 0.19  & 0.03 \\
\midrule
\multirow{4}{*}{\bagdynamic}   & \multirow{2}{*}{Alice} & \hamming & 0.01 & 0.12  & 0.03 \\
                                    &                        & \jsd   & 0.04 & 0.22   & 0.04\\
                                    \cmidrule(r){2-3}
                                    & \multirow{2}{*}{Rob}   & \hamming & 0.00 & 0.07   & 0.00\\
                                    &                        & \jsd  & 0.03 & 0.31  & 0.03 \\
\bottomrule
\end{tabular}
\caption{Results computed using the synthetic data of multi-agent teamwork, with a fully supervised training set of $200$ trajectories and known $T_x$.}
\label{table. 200 samples}
\end{table}

\balance

\end{document}